# ChatGPT and Gemini participated in the Korean College Scholastic Ability Test - Earth Science I

Seok-Hyun Ga[a,b,c], Chun-Yen Chang[a,c,d,e,1]

[a]*Institute for Research Excellence in Learning Sciences, Taipei 116, Taiwan*
[b]*The Center for Educational Research, Seoul National University, Seoul 08826, Korea*
[c]*Graduate Institute of Science Education, National Taiwan Normal University, Taipei 116, Taiwan*
[d]*Department of Earth Sciences, National Taiwan Normal University, Taipei 116, Taiwan*
[e]*Department of Biology, Universitas Negeri Malang, Malang 65145, Indonesia*

**Abstract**

The rapid development of Generative AI is bringing innovative changes to education and assessment. As the prevalence of students utilizing AI for assignments increases, concerns regarding academic integrity and the validity of assessments are growing. This study utilizes the Earth Science I section of the 2025 Korean College Scholastic Ability Test (CSAT) to deeply analyze the multimodal scientific reasoning capabilities and cognitive limitations of state-of-the-art Large Language Models (LLMs), including GPT-4o, Gemini 2.5 Flash, and Gemini 2.5 Pro. Three experimental conditions (full-page input, individual item input, and optimized multimodal input) were designed to evaluate model performance across different data structures. Quantitative results indicated that unstructured inputs led to significant performance degradation due to segmentation and Optical Character Recognition (OCR) failures. Even under optimized conditions, models exhibited fundamental reasoning flaws. Qualitative analysis revealed that "Perception Errors" were dominant, highlighting a "Perception-Cognition Gap" where models failed to interpret symbolic meanings in schematic diagrams despite recognizing visual data. Furthermore, models demonstrated a "Calculation-Conceptualization Discrepancy," successfully performing calculations while failing to apply the underlying scientific concepts, and "Process Hallucination," where models skipped visual verification in favor of plausible but unfounded background knowledge. Addressing the challenge of unauthorized AI use in coursework, this study provides actionable cues for designing "AI-resistant questions" that target these specific cognitive vulnerabilities. By exploiting AI's weaknesses, such as the gap between perception and cognition, educators can distinguish genuine student competency from AI-generated responses, thereby ensuring assessment fairness.

1. Introduction

The emergence of Generative AI technology is leading a paradigm shift across academia and society at large (O'Dea, 2024). Large Language Models (LLMs), such as OpenAI's GPT series and Google's Gemini, have learned from vast amounts of text data to demonstrate human-like language generation and comprehension abilities, expanding their influence in various fields. In particular, as they have achieved performance comparable to or surpassing human experts on standardized exams for professional fields like medicine, law, and business consulting, discussions about the intelligence level and potential of LLMs have intensified (Katz et al., 2024; Kung et al., 2023; Nori et al., 2023).

This technological achievement profoundly impacts the field of education. LLMs have raised expectations for their use as personalized tutors providing learning paths tailored to individual student levels and needs, authoring tools for automatically generating educational content, and even assessment tools for scoring and analyzing complex evaluation tasks (Evangelista, 2025; Jiang & Jiang, 2024; Liu et al., 2025; Yu et al., 2024). Recently, as models have become equipped with multimodal capabilities to process various data forms beyond text, such as images, sounds, and charts, their educational applicability has expanded even further (Chen & Wu, 2024; Denny et al., 2024; Kasneci et al., 2023; Leiker, 2023).

However, the rapid advancement of LLMs also creates significant tension, raising fundamental questions about the nature of education and assessment. On one hand, LLMs are recognized as innovative tools with immense educational potential (Bura & Myakala, 2024; Monib et al., 2024; Zapata-Rivera et al., 2024). On the other hand, the potential for misuse in assessment situations raises serious concerns about academic integrity and the fairness of evaluations (Eaton, 2022; Evangelista, 2025; Scarfe et al., 2024). How can we distinguish between a student solving a task using an LLM and solving it with their own pure ability? How can we educate learners to critically evaluate plausible but false information (hallucinations) generated by AI? These questions are urgent challenges facing educators and researchers in the AI era.

Amidst this tension, this study seeks to move beyond the perspective of viewing high-stakes assessment tests merely as a 'race' to measure LLM performance. Instead, we aim to utilize high-stakes assessment tests as a 'microscope' to deeply probe the problem-solving processes of LLMs and diagnose the cognitive characteristics of artificial intelligence. While much existing research has focused on performance-centric questions like "Can AI pass the test?", this study seeks to answer a more fundamental, process-oriented question: "How does AI attempt to solve test problems, and what does that process reveal about its cognitive structure?" Through this, we can understand the strengths and weaknesses of LLMs more precisely and, based on that understanding, explore ways to integrate AI into education more effectively and responsibly.

*1.1. The Korean CSAT as a Benchmark for Multimodal Reasoning*

A suitable benchmark is essential for evaluating the multimodal reasoning capabilities of LLMs. Existing benchmarks like MMLU (Massive Multitask Language Understanding) and SuperGLUE are useful for measuring broad knowledge, but they are mostly composed of multiple-choice questions and tend to be biased towards knowledge recall (Hendrycks et al., 2020; Wang et al., 2019). These benchmarks have limitations in assessing a model's deep reasoning processes or complex data interpretation skills, and they do not adequately reflect real-world problem-solving situations that require the organic integration of visual and textual information (Denny et al., 2024).

In this context, the Korean College Scholastic Ability Test (CSAT) provides an ideal test environment for evaluating the high-order thinking skills of LLMs (O'Dea, 2024). The CSAT is a high-stakes exam designed not just for simple knowledge memorization, but to measure the ability to think critically based on given data and to integrate multiple concepts to apply them to new problems. In particular, the Earth Science I subject in the science inquiry domain comprehensively covers various scientific concepts and principles from astronomy, geology, atmospheric science, and oceanography, and mandatorily includes various forms of visual data such as maps, graphs, charts, and schematic diagrams alongside text. Figure 1 shows a representative example from the 2025 CSAT Earth Science I section.

This question clearly demonstrates the core features of the CSAT assessment. First, it requires complex data interpretation rather than simple concept memorization. It demands a comprehensive understanding of oceanographic phenomena by simultaneously analyzing two different graphs (water temperature and salinity). Second, it requires inference of temporal change. Students must grasp annual marine environment changes through seasonal data from August, November, and February. Third, it demands the application of physical concepts. Students must understand the effects of water temperature and salinity on seawater density and apply this to the actual data. Fourth, it is characterized by multiple-choice integrated reasoning. Since the correct combination of statements (a, b, c) must be found after independently judging each one, reaching the correct answer with only partial knowledge is not possible.

Figure 1 Example of Korean CSAT problem

CSAT Earth Science I problems inherently require the fusion of heterogeneous data, which directly aligns with the core challenges faced by multimodal AI systems: data fusion and integrated reasoning capabilities (Nori et al., 2023; Yang et al., 2023). Therefore, the CSAT Earth Science I can function as a sophisticated tool for diagnosing the multimodal scientific reasoning abilities of LLMs.

2. Research Questions

This study explores the multimodal scientific reasoning capabilities of generative language models from various angles through the following research questions:

1) When an entire test paper page containing multiple questions is input as a single image, what level of performance do major LLMs show on the 2025 CSAT Earth Science I section?

2) When each question is separated and input as an individual image file, what performance do major LLMs show, and is the accuracy of question recognition and interpretation improved?

3) When text and images are separated and input in a well-structured format, what performance do major LLMs show, and is the accuracy of question recognition and interpretation improved?

4) What qualitative characteristics does the problem-solving process of LLMs, as shown in 'Research Question 3', exhibit?

## 3. Method

*3.1. Models and Experimental Environment*

This study selected Google's Gemini 2.5 Flash, Gemini 2.5 Pro, and OpenAI's GPT-4o as analysis targets (Table 1). These models are the latest versions available as of May 2025 and possess powerful multimodal processing capabilities, making them suitable for the purpose of this research. The experiments were conducted via Google AI Studio and the AI Toolkit extension in Visual Studio Code. Key parameters such as Temperature and Top P were set to the default values of each platform to observe the models' general response tendencies (O'Dea, 2024).

*Table 1* **Used model of this research**

| Model Code | Model Name |
| --- | --- |
| `gemini-2.5-flash-preview-05-20` | Gemini 2.5 Flash (Preview, 2025-05-20) |
| `gpt-4o-2024-11-20` | GPT-4o |
| `gemini-2.5-pro-preview-05-06` | Gemini 2.5 Pro (Preview, 2025-05-06) |

*3.2. Assessment Tool and Experimental Design*

The assessment tool is the Earth Science I subject from the Science Inquiry section of the '2025 College Scholastic Ability Test' administered on November 14, 2024 (Korea Institute for Curriculum and Evaluation, 2024). This test consists of 20 questions, all of which are 5-option multiple-choice. The official past question file (PDF) released by the "Korea Institute for Curriculum and Evaluation" (KICE), the institution responsible for creating and administering the test, was used. To evaluate the multimodal information processing capabilities of LLMs from various angles, three independent experiments were designed based on the input method.

- **Experiment 1 (Full Page Input):** The entire test paper (4 pages) was converted into high-quality PNG (lossless) images at 300 DPI resolution and input individually. Each page was processed to a size of approximately 3509×4963 pixels, with an average file size of 541.5 KB (365-826 KB) per page. This evaluates the model's ability to autonomously segment individual questions within a complex document layout and to process them by linking text and visual materials. It comprehensively measures item boundary recognition, text-image linking, and overall document structure comprehension in a situation where multiple items are mixed on one page.

- **Experiment 2 (Individual Item Input):** The 20 questions were individually cropped into separate image files and optimized at 300 DPI resolution for input. Individual item images were processed to an average size of 2200×1400 pixels, with an average file size of 68.65 KB (14-240 KB) per item. During the item cropping process, Adobe Photoshop was used to set precise boundaries that included both the text and visual data areas of each question. This removes the difficulty of item segmentation and allows the model to focus on its ability to integrate text and visual information within the context of a single problem.

- **Experiment 3 (Optimized Multimodal Input)**: The text of the items was manually transcribed and provided in pure text format (UTF-8 encoding), while the visual materials (graphs, charts, schematic diagrams, etc.) were individually extracted, saved as 300 DPI PNG files, and provided along with the text. The extracted visual materials were processed to an average size of 1280×1000 pixels, with an average file size of 35.85 KB (3-205 KB) per image. During the text transcription process, all mathematical symbols, Greek letters, and special characters were accurately reproduced, and the option structure (a, b, c) and choice numbers (1-5) were maintained identically to the original. This is an ideal condition to completely eliminate the possibility of OCR errors and to intensively analyze the model's core visual information interpretation and scientific reasoning abilities.

In all experiments, the same prompt was used: "Please output the problem I entered exactly as text. What is the correct answer to these problems? Also, explain how to solve them." This required the models to perform problem recognition, answer selection, and explanation of the solving process.

*3.3. Analysis Framework*

This study combined quantitative and qualitative analyses to evaluate the multimodal scientific reasoning capabilities of LLMs from multiple dimensions. Quantitative scoring measured the models' overall achievement by comparing the final answers provided in Experiments 1, 2, and 3 with the official answer key to determine the correctness of each item and calculating the total scores. Qualitative error analysis, the core of this study, was conducted to deeply investigate the cognitive characteristics of the models, focusing on the 36 incorrect response cases observed under the optimized conditions of Experiment 3. To this end, this study applied an "AI-Assisted Inductive Analysis" approach.

This methodology is based on the six-phase Thematic Analysis framework of Braun & Clarke (2006), a cornerstone of qualitative research, but modifies it into a "Human-AI Collaboration" model to suit the AI era. Recent attempts in qualitative research have sought to utilize AI not merely as an automation tool but as an analytical partner or valuable collaborator that augments the researcher's interpretation and reflection (Christou, 2024; Yan et al., 2024). Adopting this approach, this study was designed so that AI functions as "an aid in thematic analysis, enhancing the depth and breadth of analysis" rather than "overshadowing the analyst's critical evaluative and interpretive skills" (Christou, 2024, p. 560).

The specific analysis procedure underwent a four-stage process of iterative refinement.

First is the data familiarization and initial code generation phase (Braun & Clarke, 2006, Phases 1-2), where the researcher became familiar with the data by repeatedly reviewing the texts of the 36 incorrect cases. During this process, initial codes such as "visual information misreading," "concept application failure," "hasty judgment," and "calculation failure" were generated.

Second is the AI initial draft generation phase. The researcher provided the initial codes from Phase 1 and the data of the 36 incorrect cases to the AI (the Gemini model used in this study). Acting as a valuable collaborator that enhances initial data exploration and coding efficiency (Yan et al., 2024, p. 1), the AI generated a primary classification system (the initial draft of Table 2).

Third is the refinement phase through iterative feedback loops (Braun & Clarke, 2006, Phases 3-4). While the AI's initial draft from Phase 2 was useful, it revealed issues regarding trustworthiness and accuracy and limited contextual understanding as pointed out by Yan et al. (2024, p. 4). In particular, ambiguous categories such as hasty judgment encompassed multiple errors, lacking discriminatory power as an analytical framework. To address this issue, the researcher utilized the AI as a discussion partner to execute human-AI feedback loops and iterative functionality (Yan et al., 2024, p. 5). This collaborative discussion process resulted in establishing three key classification criteria:

- Through in-depth discussion on errors lumped together as hasty judgment in the initial classification, it was discovered that two fundamentally different cognitive failures were mixed. One type involved skipping core reasoning processes such as visual data verification and hastily applying related background knowledge, while the other involved setting a "False Premise" leading to logical leaps. Through this discussion, the existing reasoning errors were clearly separated and redefined into "3.1: Flawed Reasoning" and "4.2: Process Hallucination."

- It was also confirmed that visual information misreading was not a singular error. Cases of misreading quantitative values in x-y graphs and cases of failing to interpret symbolic rules in schematic diagrams or atypical figures needed to be clearly distinguished. Based on this, perception errors were subdivided into "1.1: Visual Data Misreading" and "1.2: Schematic Misinterpretation."

- In some cases, a failure type unique to AI was captured where calculation was performed correctly, but its conceptual meaning could not be connected. This was named the specific category of "2.2: Calculation-Conceptualization Gap."

Table 2 *Categorization of LLMs error*

| Error Category | Sub-category | Description | Example of Actual Error in Experiment 3 (Model & Item) |
|---|---|---|---|
| 1. Perception Errors (Perception-Cognition Gap) | 1.1: Visual Data Misreading | Misreads quantitative values (numbers, axes, trends) from x-y graphs, tables. | [Gemini 2.5 Pro (Q8)] Misread the quantitative value of the axial tilt in Period B from the graph. |
| | 1.2: Schematic Misinterpretation | Fails to understand the symbolic meaning (rules, processes) of diagrams, non-typical schematics. | [Gemini 2.5 Pro (Q13)] Failed to interpret the rules of an atypical schematic figure showing changes in typhoon wind direction. |
| 2. Conceptual Errors (Failure in Applying Conceptual Knowledge) | 2.1: Concept Misapplication | Perceived data correctly, but applied the wrong scientific principle/definition. | [Gemini 2.5 Pro (Q18)] Judged that eclipse phenomena occur at maximum blue shift (Misapplication of concepts regarding the relationship between stellar eclipses and spectral shifts). |
| | 2.2: Calculation-Concept Discrepancy | Performed the procedural calculation, but failed to connect the result to its conceptual meaning. | [GPT-4o (Q20)] Calculated $\frac{\lambda_{max,(B)}}{\lambda_{max,(C)}} = 0.5$ correctly but failed to connect this to the concept that $\lambda_{max,(C)}$ is twice as large. |
| 3. Reasoning Errors (Logical Inconsistency) | 3.1: Flawed Reasoning | Makes a logical leap or sets a false premise to reach a wrong conclusion. | [Gemini 2.5 Pro (Q17)] Skipped the complex identification process based on data (table) and established a baseless premise (e.g., assuming A is dark energy) to solve the problem. |
| | 3.2: Spatio-temporal Failure | Fails to reconstruct dynamic changes over time or 3D spatial relationships. | [GPT-4o (Q4)] Misunderstood the relationship between the hot spot (fixed) and the plate (moving), leading to an incorrect deduction of the plate's direction of movement opposite to the volcanic island arrangement (A→E). |
| 4. Knowledge/Generation Errors (Hallucination) | 4.1: Factual Hallucination | Presents irrelevant or scientifically incorrect external knowledge as a basis. | [GPT-4o (Q7)] Presented inaccurate external knowledge stating that "fusulinids" (which went extinct at the end of the Permian) went extinct during Period A (end of the Ordovician). |
| | 4.2: Process Hallucination | Skips a complex reasoning step (e.g., visual verification) and substitutes a plausible related background knowledge. | [GPT-4o (Q3)] Skipped the visual evidence verification (reasoning process) of the P-T diagram and hastily applied related background knowledge called decompression melting without data verification. |

Finally, there is the human-led verification and finalization phase (Braun & Clarke, 2006 Phase 5). The refined analysis framework of 4 major categories and 8 sub-categories (Table 2), derived from the collaborative discussion in Phase 3, was finalized. This process is a human-led stage where the researcher does not blindly accept the AI's suggestions but exercises final critical evaluative and interpretive skills (Christou, 2024, p. 560) to thoroughly and meticulously evaluate and validate the AI-generated results (Christou, 2024, p. 567). Finally, to ensure the reliability of this classification scheme, inter-coder reliability was secured by requesting peer validation of the final classification system and original data from a colleague with expertise in AI education and qualitative research.

## 4. Result & Discussion

The core of this research lies not just in whether the LLMs got the answers right, but in analyzing how they solve the problems. The experimental results reveal that the models exhibit complex cognitive characteristics that cannot be grasped by a simple binary evaluation of correct/incorrect. Notably, problem-solving abilities showed dramatic differences depending on the structural level of the input information, and fundamental limitations in scientific reasoning were observed even under optimized conditions.

Specific differences in cognitive characteristics were observed among the models. GPT-4o, while relatively maintaining logical consistency, had the lowest OCR accuracy and frequent hallucinations. Gemini 2.5 Flash showed systematic errors in visual information interpretation but was relatively accurate in calculation processes. Gemini 2.5 Pro demonstrated the most balanced performance overall but still revealed consistent limitations in problems requiring complex reasoning.

*4.1. Research Question 1: Overwhelming Challenges in Full-Page Input*

The results of Experiment 1 (Full Page Input) clearly show that current LLMs have fundamental limitations in processing complex and unstructured visual documents. In Experiment 1, Gemini 2.5 Flash scored 4 points (8%), GPT-4o scored 7 points (14%), and Gemini 2.5 Pro scored 14 points (28%) (Table 3). Compared to the expected value of random choice (10 points, 20%), Gemini 2.5 Flash and GPT-4o performed worse than random, and even Gemini 2.5 Pro struggled with basic problem recognition. This implies that the models failed even at the problem recognition stage, before the problem-solving stage, demonstrating that the multimodal information processing pipeline of LLMs is very brittle.

Particularly noteworthy are the systematic failures that occurred during the OCR process. All models struggled to accurately recognize key visual information or text in the questions, and these recognition failures led to severe conceptual misunderstandings. In question 5, GPT-4o misidentified a problem about a galaxy's spectrum as a meteorite spectrum problem, rendering all subsequent reasoning about celestial classification and spectral analysis meaningless. Gemini 2.5 Flash showed an even more severe error; in question 3, it completely misinterpreted a problem about a rock's melting curve and magma as a problem about a rock salt melting curve and non-existent terms like 'fluidal rock salt' and 'double salt'.

Both models exhibited 'hallucination' phenomena, generating plausible scientific explanations based on such incorrect problem recognition. For example, in question 17 (cosmology), GPT-4o inversely interpreted the change in component densities but still mentioned the correct knowledge that 'the proportion of dark energy increases over time'. Similarly, Gemini 2.5 Flash provided a plausible (though irrelevant to the problem) explanation for its 'rock salt' problem in question 3, stating that 'rock salt can melt as temperature rises'. This is particularly dangerous in an educational context, as learners may trust the model's confident but incorrect explanations.

*Table 3* **Results of Experiment 1**

| Item No. | Answer | Score | Gemini 2.5 Flash | GPT-4o | Gemini 2.5 Pro Pre |
|---|---|---|---|---|---|
| 1 | ④ | 2 | X (⑤) | X (⑦) | ✓ (④) |
| 2 | ⑤ | 2 | X (④) | ✓ (⑤) | ✓ (⑤) |
| 3 | ④ | 2 | ✓ (④) | X (⑤) | X (⑤) |
| 4 | ③ | 3 | X (②) | X (⑥) | ✓ (③) |
| 5 | ④ | 2 | X (②) | X (⑤) | ✓ (④) |
| 6 | ② | 2 | X (④) | ✓ (②) | X (⑤) |
| 7 | ② | 2 | X (①) | X (①) | X (④) |
| 8 | ③ | 2 | X (⑤) | X (⑤) | X (⑤) |
| 9 | ① | 3 | X (④) | ✓ (①) | X (④) |
| 10 | ① | 3 | X (②) | X (③) | X (②) |
| 11 | ③ | 3 | X (④) | X (②) | X (①) |
| 12 | ⑤ | 2 | X (①) | X (①) | X (②) |
| 13 | ② | 3 | X (④) | X (⑤) | X (④) |
| 14 | ⑤ | 3 | X (②) | X (④) | X (②) |
| 15 | ① | 2 | X (②) | X (④) | ✓ (①) |
| 16 | ⑤ | 3 | X (①) | X (④) | X (③) |
| 17 | ⑤ | 3 | X (②) | X (①) | ✓ (⑤) |
| 18 | ② | 3 | X (⑤) | X (④) | X (①) |
| 19 | ④ | 3 | X (①) | X (③) | X (②) |
| 20 | ② | 2 | ✓ (②) | X (①) | X (④) |
| Total Score | | 50 | 4 points | 7 points | 14 points |
| Accuracy | | | 8% | 14% | 28% |

*4.2. Research Question 2: Uneven Performance Improvement in Individual Item Input and Its Causes*

In Experiment 2 (Individual Item Input), providing items separately led to dramatic differences in performance improvement among models. Gemini 2.5 Flash scored 9 points (18%), GPT-4o scored 8 points (16%), while Gemini 2.5 Pro scored 28 points (56%) (Table 4). Interestingly, Gemini 2.5

Flash and GPT-4o showed only minimal improvement compared to Experiment 1 (4 and 7 points, respectively), whereas only Gemini 2.5 Pro showed significant improvement, from 14 to 28 points.

This stark performance difference clearly reveals the core bottleneck of each model. The superior performance improvement of Gemini 2.5 Pro suggests its main failure cause was not a 'lack of scientific reasoning ability' but rather the 'complex document layout analysis and item segmentation failure' of Experiment 1. In other words, once the input data was structured into individual items, the interference problem caused by the mixing of multiple items was resolved, and the model's inherent (and relatively superior) multimodal reasoning ability could be exerted.

Conversely, the minimal improvement of Gemini 2.5 Flash and GPT-4o implies their limitations are more fundamental (in the reasoning ability itself). This reveals the limits of partial improvement even when layout issues are resolved. This analysis is supported by the specific error patterns in Experiment 2.

First, OCR errors in recognizing key visual information within the item still occurred. This shows the OCR problem was only partially resolved. In question 1, both Gemini 2.5 Flash and GPT-4o made the same error of inversely identifying ripple marks (actual answer C) and mud cracks (actual answer A). In question 5, they also showed clear visual information misreading, misidentifying the absorption line spectrum of an elliptical galaxy (a) as a 'continuous spectrum' or 'Seyfert galaxy', and the emission line spectrum of a Seyfert galaxy (b) as 'absorption lines' or 'elliptical galaxy'.

Second, a gap between Perception and Cognition was found. That is, even if they recognized the text or visual information, the reasoning ability to connect it to the correct scientific concept was lacking. For example, in question 1, GPT-4o, after judging (albeit misreading) A as 'mud cracks' and C as 'ripple marks', reached the incorrect answer (5) through contradictory reasoning by also judging statement (c) 'There was a period of exposure to a dry environment while C (ripple marks) was forming' as correct. This shows a failure in scientific reasoning, regardless of OCR accuracy (or inaccuracy).

Third, consistent vulnerabilities were observed. Particularly in questions 4 (hotspot experiment), 9 (atmospheric general circulation), 13 (typhoon), and 19 (paleomagnetism), all three models chose incorrect answers. The common feature of these questions is that they require complex spatio-temporal reasoning or the integrated application of multiple scientific principles. For example, in the hotspot experiment (question 4), Gemini 2.5 Flash failed in basic geological concept understanding, misclassifying the fixed point (a) as a simple 'reference point' instead of a 'hotspot'. This indicates that certain types of high-order reasoning remain a challenge for current models, regardless of input format improvements.

*Table 4* **Results of Experiment 2**

| Item No. | Answer | Score | Gemini 2.5 Flash | GPT-4o | Gemini 2.5 Pro Pre |
|---|---|---|---|---|---|
| 1 | ④ | 2 | X (⑤) | X (⑤) | ✓ (④) |
| 2 | ⑤ | 2 | ✓ (⑤) | ✓ (⑤) | ✓ (⑤) |
| 3 | ④ | 2 | ✓ (④) | X (⑤) | ✓ (④) |
| 4 | ③ | 3 | X (④) | X (⑤) | X (⑤) |
| 5 | ④ | 2 | X (⑤) | X (①) | ✓ (④) |
| 6 | ② | 2 | X (④) | X (①) | X (⑤) |
| 7 | ② | 2 | X (④) | X (④) | X (④) |
| 8 | ③ | 2 | X (⑤) | X (④) | ✓ (③) |
| 9 | ① | 3 | X (⑤) | X (⑤) | X (⑤) |
| 10 | ① | 3 | X (④) | X (④) | X (②) |
| 11 | ③ | 3 | X (⑤) | X (④) | ✓ (③) |
| 12 | ⑤ | 2 | X (④) | X (①) | ✓ (⑤) |
| 13 | ② | 3 | X (④) | X (④) | X (④) |
| 14 | ⑤ | 3 | ✓ (⑤) | ✓ (⑤) | ✓ (⑤) |
| 15 | ① | 2 | ✓ (①) | X (⑤) | ✓ (①) |
| 16 | ⑤ | 3 | X (④) | ✓ (⑤) | X (①) |
| 17 | ⑤ | 3 | X (②) | X (④) | ✓ (⑤) |
| 18 | ② | 3 | X (④) | X (④) | ✓ (②) |
| 19 | ④ | 3 | X (⑤) | X (③) | X (②) |
| 20 | ② | 2 | X (④) | X (④) | ✓ (②) |
| Total Score | | 50 | 9 points | 8 points | 28 points |
| Accuracy | | | 18% | 16% | 56% |

*4.3. Research Question 3: Subtle Limits of Multimodal Reasoning Under Optimized Conditions*

In Experiment 3, OCR and item segmentation problems were artificially eliminated to evaluate the models' pure reasoning abilities. As a result, Gemini 2.5 Flash scored 10 points (20%), GPT-4o scored 11 points (22%), and Gemini 2.5 Pro scored 34 points (68%) (Table 5). This shows that while all models' performance improved compared to Experiments 1 and 2, the performance gap between models remained stark. It is particularly noteworthy that while Gemini 2.5 Pro achieved a performance level approaching that of high-performing examinees, the other two models remained at the level of random choice (20%). This suggests the fundamental impact of model architecture and training data on scientific reasoning capabilities. However, even under these ideal conditions, all models showed systematic error patterns, clearly indicating that the limitations of current LLMs lie in the depth of conceptual understanding and reasoning, beyond simple information processing capabilities.

*Table 5* **Results of Experiment 3**

| Item No. | Answer | Score | Gemini 2.5 Flash | GPT-4o | Gemini 2.5 Pro |
|---|---|---|---|---|---|
| 1 | ④ | 2 | ✓ (④) | ✓ (④) | ✓ (④) |
| 2 | ⑤ | 2 | ✓ (⑤) | ✓ (⑤) | ✓ (⑤) |
| 3 | ④ | 2 | X (⑤) | X (⑤) | ✓ (④) |
| 4 | ③ | 3 | X (⑤) | X (⑤) | ✓ (③) |
| 5 | ④ | 2 | ✓ (④) | X (⑤) | ✓ (④) |
| 6 | ② | 2 | X (⑤) | X (④) | X (⑤) |
| 7 | ② | 2 | ✓ (②) | X (④) | ✓ (②) |
| 8 | ③ | 2 | X (⑤) | X (⑤) | X (①) |
| 9 | ① | 3 | X (⑤) | X (⑤) | X (②) |
| 10 | ① | 3 | X (④) | ✓ (①) | ✓ (①) |
| 11 | ③ | 3 | X (④) | X (⑤) | ✓ (③) |
| 12 | ⑤ | 2 | X (②) | ✓ (⑤) | ✓ (⑤) |
| 13 | ② | 3 | X (⑤) | X (④) | X (④) |
| 14 | ⑤ | 3 | X (④) | X (④) | ✓ (⑤) |
| 15 | ① | 2 | ✓ (①) | ✓ (①) | ✓ (①) |
| 16 | ⑤ | 3 | X (②) | X (④) | ✓ (⑤) |
| 17 | ⑤ | 3 | X (④) | X (②) | X (①) |
| 18 | ② | 3 | X (④) | X (④) | X (⑤) |
| 19 | ④ | 3 | X (⑤) | X (⑤) | ✓ (④) |
| 20 | ② | 2 | X (③) | X (③) | ✓ (②) |
| Total Score | | 50 | 10 points | 11 points | 34 points |
| Accuracy | | | 20% | 22% | 68% |

*4.4. Research Question 4: Qualitative Analysis of LLMs' Reasoning Processes*

The core of this research lies not just in whether the LLMs got the answers right, but in analyzing how they solve the problems. The results of Experiment 3 reveal that the models exhibit complex cognitive characteristics that cannot be grasped by a simple binary evaluation of correct/incorrect. Notably, fundamental limitations in scientific reasoning were observed even under ideal, optimized input conditions. This suggests that current LLM technology still has limitations in deep conceptual understanding and flexible reasoning abilities, beyond surface-level information processing.

To deeply analyze the LLM problem-solving process, this study proposes a new framework for systematically classifying AI errors. This framework divides the problem-solving process into four stages, reflecting human cognitive processes: Perception, Conceptualization, Reasoning, and Knowledge and Generation, and defines the error types that can occur at each stage. This aims to precisely diagnose at which point in the AI's cognitive pipeline the failure occurs, rather than just listing errors.

*4.4.1. Statistical Analysis of Error Causes*

A total of 36 incorrect answer cases (15 from Gemini 2.5 Flash, 15 from GPT-4o, and 6 from Gemini 2.5 Pro) generated in Experiment 3 were deeply analyzed according to the classification table in Table 2. In this process, when a single incorrect answer involved two errors simultaneously, each error was counted as 0.5. There were no cases with three or more errors. As a result, a total of 36 individual error types were tallied.

The re-aggregated statistics (Table 6) provide a new perspective on the core vulnerabilities of LLMs. The most prominent result is that "Perception Errors" emerged as the most dominant error type, accounting for a total of 43.06% (15.5 cases). This implies that LLMs experience significant difficulties from the fundamental perception stage—whether reading quantitative data from given visual materials (1.1: 9 cases) or interpreting symbolic meanings (1.2: 6.5 cases)—before even attempting high-order reasoning. Next, "Reasoning Errors" accounted for 25% (9 cases), including failures in logical leaps (3.1) or spatiotemporal reasoning (3.2). "Knowledge/Generation Errors" (16.67%, 6 cases) and "Conceptual Errors" (15.28%, 5.5 cases) followed.

Model-specific characteristics were also clearly revealed. Both Gemini 2.5 Flash and GPT-4o exhibited the highest number of error types (15 cases each). However, GPT-4o showed a relatively even distribution across "Perception Errors" (5.5 cases), "Knowledge/Generation Errors" (4 cases), and "Conceptual Errors" (3.5 cases), showing complex errors in various areas. Notably, regarding "Knowledge/Generation Errors" (4.1, 4.2), GPT-4o accounted for 4 cases, while Gemini 2.5 Flash accounted for 2 cases. Specifically, all 2 cases of "Factual Hallucination" (4.1) occurred exclusively in GPT-4o, whereas "Process Hallucination" (4.2) was observed in both Gemini 2.5 Flash (2 cases) and GPT-4o (2 cases). On the other hand, the errors of Gemini 2.5 Pro, which showed the highest performance (total 6 cases), were overwhelmingly concentrated in "Perception Errors" (4 cases) and "Reasoning Errors" (1.5 cases), clearly demonstrating where the current frontier of AI scientific reasoning capability lies.

Table 6 *Statistics on the Number of Reasons for Errors*

| Main Error Category | Sub-category | Gemini 2.5 Flash (Count) | GPT-4o (Count) | Gemini 2.5 Pro (Count) | Total (Count) | Percentage (%) |
|---|---|---|---|---|---|---|
| 1. Perception Errors | 1.1: Visual Data Misreading | 3 | 3.5 | 2.5 | 9 | 25.00% |
| | 1.2: Schematic Misinterpretation | 3 | 2 | 1.5 | 6.5 | 18.06% |
| 2. Conceptual Errors | 2.1: Concept Misapplication | 1.5 | 2.5 | 0.5 | 4.5 | 12.50% |
| | 2.2: Calculation-Concept Gap | 0 | 1 | 0 | 1 | 2.78% |
| 3. Reasoning Errors | 3.1: Flawed Reasoning | 4.5 | 1 | 1.5 | 7 | 19.44% |
| | 3.2: Spatio-temporal Failure | 1 | 1 | 0 | 2 | 5.56% |
| 4. Knowledge /Gen. Errors | 4.1: Factual Hallucination | 0 | 2 | 0 | 2 | 5.56% |
| | 4.2: Process Hallucination | 2 | 2 | 0 | 4 | 11.11% |
| Total Error Counts | | 15 | 15 | 6 | 36 | 100.00% |

### 4.4.2. Gemini 2.5 Flash Error Analysis

Gemini 2.5 Flash recorded a total of 15 incorrect answers (16 error counts), with errors mainly concentrated in "Perception Errors" (6 cases) and "Reasoning Errors" (5.5 cases). Specifically, "Schematic Misinterpretation" (1.2), which involves failing to interpret the rules of atypical diagrams, and "Flawed Reasoning" (3.1), based on logical leaps or incorrect premises, occurred most frequently with 4 and 5 cases, respectively (See Table 7 for details).

For example, in questions 13 (Typhoon) and 18 (Planetary Orbit), "Schematic Misinterpretation" (1.2) was observed in interpreting atypical radial charts or geometric schematic diagrams. Also, in question 17, it established a baseless premise (3.1) that "C must be dark energy" without data analysis, and in question 20, it hastily concluded (3.1) that high temperature and low luminosity means impossible to be a main sequence star, revealing fundamental flaws in the logical reasoning process. This shows the model's vulnerability in grasping the meaning of complex visual information and constructing solid logic based on it.

Table 7 *Detailed Analysis of the Reasons for Errors in Gemini 2.5 Flash*

| Item No. | Score | Answer | AI Response | Error Category | Detailed Analysis |
|---|---|---|---|---|---|
| 3 | 2 | ④ | ⑤ | 4.2: Process Hallucination | Avoided the core reasoning process of visually verifying whether the melting curve is crossed during decompression (vertical rise) at point ⓑ in the P-T diagram (Figure). Hastily applied related background knowledge called decompression melting without data verification. |
| 4 | 3 | ③ | ⑤ | 3.2: Spatio-temporal Failure | Failed in spatiotemporal/dynamic reasoning to deduce the plate's movement direction (northwest) through the volcanic island arrangement order (A→E) and formation timing based on the relationship between the hot spot (fixed) and the plate (moving). |
| 6 | 2 | ② | ⑤ | 2.1: Concept Misapplication | (2.1) Due to a lack of understanding of the 3D structure of extratropical cyclones, incorrectly judged that a warm front surface still exists above point A after the warm front has passed. |
| | | | | 1.2: Schematic Misinterpretation | (1.2) Also, failed to recognize the core interpretation rule that brightness in the visible satellite image of Figure (B) symbolizes cloud thickness/amount. (Consequently ignored visual data and used inaccurate grounds of relative position). |
| 8 | 2 | ③ | ⑤ | 1.1: Visual Data Misreading | Repeatedly misread quantitative data such as eccentricity and axial tilt for periods A and B in the data graph (a). |
| 9 | 3 | ① | ⑤ | 1.1: Visual Data Misreading | Misread the quantitative data (45°) of point B as 60° on the horizontal axis (latitude) of the data graph. |
| 10 | 3 | ① | ④ | 3.1: Flawed Reasoning | Showed contradictory and leaping causal reasoning by proceeding with calculations under the premise of equal luminosity in option (b), despite having already confirmed different luminosity in option (a). |
| 11 | 3 | ③ | ④ | 1.1: Visual Data Misreading | Misread the quantitative data that the y-values (age) of (a) and (b) are the same in the data graph (B), and also misread the x-values (distance) data of (c) and (d). |
| 12 | 2 | ⑤ | ② | 1.2: Schematic Misinterpretation | Failed to interpret the scientific rule (clockwise/counter-clockwise change in wind direction) symbolized by the atypical radial schematic figure showing changes in typhoon wind direction/pressure. |

| Item No. | Score | Answer | AI Response | Error Category | Detailed Analysis |
|---|---|---|---|---|---|
| 13 | 3 | ② | ⑤ | 2.1: Concept Misapplication | Failed to apply scientific concepts regarding the difference in core physical quantities (core temperature) according to the evolutionary stages of the giant star (a) and main sequence star (b) on the H-R diagram. |
| 14 | 3 | ⑤ | ④ | 3.1: Flawed Reasoning | Performed reasoning with severe logical leaps, relying on guessing without clearly calculating the half-life of Y, and concluding the change in Y content approximately without specific calculation. |
| 16 | 3 | ⑤ | ② | 3.1: Flawed Reasoning | Skipped the identification process based on data (table) and proceeded with reasoning by setting a baseless premise (False Premise) from the beginning that 'C must be dark energy'. |
| 17 | 3 | ⑤ | ④ | 1.2: Schematic Misinterpretation | Misread the geometric relationship (angle) of the atypical schematic diagram (a) showing planetary orbit and phase angle. |
| 18 | 3 | ② | ④ | 1.2: Schematic Misinterpretation | (1.2) Failed to recognize the core rule of the diagram that the y-axis of the magnetic inclination graph (A) includes not only 'magnitude' but also 'direction' (+/-). |
|   |   |   |   | 3.1: Flawed Reasoning | (3.1) Consequently failed in causal reasoning for calculating the amount of latitude change. |
| 19 | 3 | ④ | ⑤ | 3.1: Flawed Reasoning | Hastily concluded 'impossible to be a main sequence star' (result) from 'high surface temperature and low luminosity' (cause) and misjudged it as a white dwarf. (Failure in causal reasoning considering position on the H-R diagram). |
| 20 | 2 | ② | ③ | 4.2: Process Hallucination | Instead of performing relative calculations based on data, misjudged the absolute magnitude of (A) by forcibly applying general background knowledge that 'white dwarfs are very faint'. |

### 4.4.3. OpenAI GPT-4o Error Analysis

GPT-4o recorded 15 incorrect answers (20 total error instances), exhibiting a complex error pattern with a relatively even distribution across "Perception Errors" (8 cases), "Conceptual Errors" (5 cases), and "Knowledge/Generation Errors" (5 cases). Notably, the most distinctive feature of GPT-4o was the "Hallucination" phenomenon. In Question 7, it used inaccurate external knowledge (4.1) about "fusulinids" (extinct at the end of the Permian), which was irrelevant to Period A (end of the Ordovician), as evidence. A more severe type was "Process Hallucination" (4.2) observed in Question 3. This question required "visual verification" of a P-T diagram, but GPT-4o "Skipped" this core reasoning process. Instead, it activated related background knowledge of "decompression melting" in the context of "mantle rising" and hastily applied it without data verification to justify its conclusion. This demonstrates the deceptive nature of AI attempting to evade difficult reasoning and generate plausible answers (See Table 8 for details).

*Table 8 Detailed Analysis of the Reasons for Errors in OpenAI GPT-4o*

| Item No. | Score | Answer | AI Response | Error Category | Detailed Analysis |
|---|---|---|---|---|---|
| 3 | 2 | ④ | ⑤ | 4.2: Process Hallucination | Avoided visual evidence verification (core reasoning process) of the P-T diagram and hastily applied related background knowledge called decompression melting without data verification to justify the conclusion. |
| 4 | 3 | ③ | ⑤ | 3.2: Spatio-temporal Failure | (Same as Flash) Failed in the dynamic process (spatiotemporal reasoning) of interpreting the spatiotemporal meaning of volcanic island arrangement to infer the direction of plate movement. |
| 5 | 2 | ④ | ⑤ | 2.1: Concept Misapplication | Recognized the visual features of the spectra (A, B) but committed a concept misapplication by connecting them inversely to Seyfert galaxies and elliptical galaxies, respectively. |
| 6 | 2 | ② | ④ | 2.1: Concept Misapplication | (2.1) Applied the incorrect scientific concept that wind direction changes counter-clockwise when a warm front passes. |
| | | | | 1.1: Visual Data Misreading | (1.1) Also, misread the visual data of contrast (brightness) between (a) and (b) in the satellite image (B) inversely. |
| 7 | 2 | ② | ④ | 4.1: Factual Hallucination | Brought in inaccurate external knowledge irrelevant to the question, stating that 'fusulinids' (extinct at the end of the Permian) went extinct, as evidence for the mass extinction in Period A (end of the Ordovician). (Factual Hallucination) |
| 8 | 2 | ③ | ⑤ | 1.1: Visual Data Misreading | Showed total data misreading by perceiving the axial tilt and eccentricity curves inversely and misreading quantitative values on the graph. |
| 9 | 3 | ① | ⑤ | 4.2: Process Hallucination | (4.2) Exhibited 'Process Hallucination' by assuming the location of A is 30°N without any basis when there was no information about latitude. |
| | | | | 1.1: Visual Data Misreading | (1.1) Also misread the latitude of B as the equator (0°). |
| 11 | 3 | ③ | ⑤ | 1.2: Schematic Misinterpretation | (1.2) Committed schematic misinterpretation by failing to recognize that (a) and (b) belong to different plates in the plate boundary schematic (A). |
| | | | | 2.2: Calculation-Concept Gap | (2.2) Failed to connect the concept that relative velocity is the sum of two velocities to the calculation. |
| 13 | 3 | ② | ④ | 1.2: Schematic Misinterpretation | (Same as Flash) Failed to interpret the scientific rules symbolized by the atypical radial schematic figure showing wind direction/pressure changes. |
| 14 | 3 | ⑤ | ④ | 2.1: Concept Misapplication | (Same as Flash) Failed to apply concepts to understand the difference in core physical quantities (core temperature) according to the positions of the giant star (a) and main sequence star (b) on the H-R diagram. |
| 16 | 3 | ⑤ | ④ | 4.1: Factual Hallucination | Generated and used scientifically completely incorrect external knowledge (Factual Hallucination) as evidence, stating that 'Y has no separate half-life because it is produced by the decay of X'. |
| 17 | 3 | ⑤ | ② | 3.1: Flawed Reasoning | Ignored the instruction ("in no particular order") and started reasoning by arbitrarily assuming (baseless premise) that A, B, and C are ordinary matter, dark matter, and dark energy, respectively. |
| 18 | 3 | ② | ④ | 1.1: Visual Data Misreading | Misread the data trend (converging to $\lambda_0$) in the interval $t_0 + 2.5T \sim t_0 + 3T$ on the data graph (B). |

| Item No. | Score | Answer | AI Response | Error Category | Detailed Analysis |
|---|---|---|---|---|---|
| 19 | 3 | ④ | ⑤ | 1.2: Schematic Misinterpretation | (1.2) Failed to recognize the rule of the diagram that the y-axis of the magnetic inclination graph (A) includes direction (+/-). |
| | | | | 1.1: Visual Data Misreading | (1.1) Also failed to accurately convert the quantitative value (40° → approx. 30°) in the inclination-latitude conversion graph (B). |
| 20 | 2 | ② | ③ | 2.2: Calculation-Concept Gap | (2.2) Knew the $\lambda_{max} \propto \frac{1}{T}$ relationship (calculation), but failed to connect the calculation result of $\frac{\lambda_{max,(B)}}{\lambda_{max,(C)}} = 0.5$ to the concept that $\lambda_{max,(C)}$ is twice as large. |
| | | | | 4.2: Process Hallucination | (4.2) Avoided complex calculations for options b and c and concluded with baseless hasty judgments. |

### 4.4.4. Gemini 2.5 Pro Error Analysis

Gemini 2.5 Pro recorded only 6 incorrect answers, demonstrating performance that overwhelmed the other two models. The errors of this model (total 8 instances) occurred mainly in high-order problems, rather than basic conceptual errors or hallucinations. Specifically, they were concentrated in "Perception Errors" (5 cases) and "Flawed Reasoning" (2 cases).

This clearly shows where the limits of the current state-of-the-art AI scientific reasoning capabilities lie. Question 9 is a representative case of "Visual Data Misreading" (1.1). This problem presented east-west wind speed and north-south wind speed as two separate graphs ((ⓐ), (ⓑ)), but the model failed at the first stage (perception) by totally misreading the legend (thin line/thick line) and axis information (latitude) of these graphs.

Question 17 is an example of "Flawed Reasoning" (3.1). The model correctly understood individual concepts such as the principles of density change in cosmic components. However, it committed logical errors in the multi-step complex reasoning process by applying these concepts to the complex numerical data in the given table and setting "baseless premises", such as "assuming A is dark energy." This shows that while AI possesses individual knowledge, it still has limitations in the ability to integrate it to solve complex problems (See Table 9 for details).

*Table 9 Detailed Analysis of the Reasons for Errors in Gemini 2.5 Pro*

| Item No. | Score | Answer | AI Response | Error Category | Detailed Analysis |
|---|---|---|---|---|---|
| 6 | 2 | ② | ⑤ | 3.1: Flawed Reasoning | (3.1) Despite starting from the correct premise that 'A is in the warm sector', committed a logical leap (failure in causal reasoning) stating that 'a cold front will pass'. |
| | | | | 1.1: Visual Data Misreading | (1.1) Also misread the brightness difference (visual data) between the bright cloud of (a) and the dark sea area of (b) in the satellite image (B) inversely. |

| Item No. | Score | Answer | AI Response | Error Category | Detailed Analysis |
|---|---|---|---|---|---|
| 8 | 2 | ③ | ① | 1.1: Visual Data Misreading | Misread the quantitative data of the axial tilt value (approx. 22.5°) for Period B in the data graph. |
| 9 | 3 | ① | ② | 1.1: Visual Data Misreading | Totally misread core quantitative and qualitative data such as the graph legend (a=thin line, b=thick line) and axis information (A=Southern Hemisphere). |
| 13 | 3 | ② | ④ | 1.2: Schematic Misinterpretation | (Same as Flash) Failed to interpret the scientific rules (clockwise/counter-clockwise change in wind direction) symbolized by the atypical radial schematic figure showing wind direction/pressure changes. |
| 17 | 3 | ⑤ | ① | 3.1: Flawed Reasoning | Understood individual concepts (density change principles) but failed in multi-step reasoning to apply them to complex data (table) to identify A, B, and C. Specifically, established baseless premises (failure in causal reasoning) such as 'thinking A is dark energy' to solve the problem. |
| 18 | 3 | ② | ⑤ | 1.2: Schematic Misinterpretation | (1.2) Misread the geometric relationship (angle) of the schematic diagram (a) showing planetary orbit and line-of-sight direction. |
|   |   |   |   | 2.1: Concept Misapplication | (2.1) Also misapplied core scientific concepts regarding the relationship between the observed object (star) and the phenomenon (eclipse), such as judging that 'eclipse phenomena occur at maximum blue shift'. |

## 5. Conclusion

This study provided an in-depth analysis of the process by which state-of-the-art Large Language Models (LLMs) solve complex multimodal scientific assessment tasks, specifically the Earth Science I section of the 2025 College Scholastic Ability Test (CSAT) in South Korea. The core objective of this study was to move beyond the mere success or failure of AI to elucidate the cognitive characteristics and current technological limitations revealed during the problem-solving process. The results clearly demonstrated that model performance is extremely sensitive to the format and structure of input information and that fundamental limitations in scientific reasoning persist even under optimized conditions.

First, the models exhibited fundamental vulnerabilities from the "Perception" stage, preceding high-order reasoning. When processing unstructured full-page exam images (Experiment 1), the models failed to analyze complex document layouts and recognize question boundaries, resulting in catastrophic OCR errors. However, a more critical issue was that even under optimized conditions where OCR errors were completely eliminated (Experiment 3), "Perception Errors" remained the dominant cause of failure, accounting for 43.06% of total errors. This suggests that AI experiences systematic failure not only in "Visual Data Misreading" (1.1) of quantitative data but also in "Schematic Misinterpretation" (1.2), failing to interpret the scientific rules symbolized by atypical diagrams or schematic figures.

Second, the models failed to integrate superficial information processing with deep conceptual understanding. This limitation was observed through two major gaps. One is the "Perception-Cognition Gap," where the failure to interpret the symbolic meaning of diagrams indicates a severed cognitive bridge connecting recognized information to scientific concepts, rather than simple visual errors. The other is the "Calculation-Conceptualization Discrepancy," where models successfully performed procedural tasks such as calculation but failed to apply the core scientific concepts implied by the calculation results.

Third, the models tended to skip complex reasoning and justify their conclusions through hallucination. The reliance of current LLMs on sophisticated pattern matching rather than genuine reasoning was evident in the hallucination types. Notably, "Process Hallucination (4.2)" was significant; the models skipped complex reasoning processes requiring visual verification of data. Instead, they hastily applied related background knowledge without data verification to justify plausible conclusions. Additionally, as seen in "Flawed Reasoning (3.1)," models sometimes injected baseless premises into the logical process to bypass complex identification steps.

These phenomena—the gap between perception and cognition, the discrepancy between calculation and concepts, and the avoidance of reasoning—suggest that current LLMs have not yet reached the stage of constructing meaning based on flexible conceptual schemas like humans. Therefore, when utilizing LLMs in educational settings, it is essential to critically review the fundamental perception errors, conceptual disconnections, and potential for reasoning hallucinations hidden behind the confident answers generated by the models. Education that fosters critical thinking skills, enabling students to verify AI responses based on scientific principles and data rather than blindly trusting them, must accompany AI adoption.

## 6. Implication and Further Research

This study has limitations in that it analyzed three specific LLMs and a single exam subject. To address these limitations and deepen the understanding of AI's educational utility, the findings present several important implications and directions for future research.

The most significant implication concerns the design of assessment items in the AI era. The clear vulnerabilities of LLMs revealed in this study—specifically the "Perception-Cognition Gap" and "Calculation-Conceptualization Discrepancy"—paradoxically suggest a direction for developing AI-resistant assessment items that target the fundamental limitations of AI, moving beyond information retrieval or simple calculation tasks that AI solves easily. For instance, educators and assessment experts can actively utilize atypical "Schematic Misinterpretation (1.2)" items, such as Questions 13 or 18, which require interpreting scientific rules symbolized by signs or shapes beyond simple x-y graphs, to evaluate the "Perception-Cognition Gap". Additionally, multi-step items can be designed where procedural calculation is only the first step and the scientific meaning of the calculation result (2.2) must be reapplied to reach the solution, as in Question 20, to target the "Calculation-Conceptualization Discrepancy". Furthermore, as seen in Question 3, items can be designed such that hasty application of background knowledge (e.g., decompression melting) leads to incorrect answers, and the correct answer can only be reached through strict verification (4.2) of the visual data, thereby preventing "Process Hallucination". Such sophisticated item design will contribute to preventing the misuse of AI and ensuring academic integrity.

The specific findings of this study call for systematic future research. First, it is necessary to verify whether the characteristic error patterns of AI found in this study, such as the "Perception-Cognition Gap" and "Calculation-Conceptualization Discrepancy," appear universally in other science subjects (Physics, Chemistry, Biology) and across a wider range of LLMs. This will contribute to defining the fundamental limitations of current AI technology.

Next, research is required to systematically verify whether advanced prompting techniques such as "Chain of Thought" or "Self-reflection" can significantly mitigate the specific errors observed in this study, particularly "Process Hallucination (4.2)" or "Flawed Reasoning (3.1)".

Finally, a critical direction for research is the direct comparative analysis between representative misconception types exhibited by human students and the LLM error types classified in this study (Table 2). For example, it is necessary to precisely determine whether human students also exhibit the "Calculation-Conceptualization Discrepancy (2.2)" like AI, or if they fail in different ways. Such comparative analysis will provide essential foundational data for designing next-generation AI-based educational systems that complement human learning processes and provide personalized feedback.

Acknowledgement

This work was supported by the National Research Foundation of Korea (Grant No. 2023S1A5B5A16080581), the National Science and Technology Council, Taiwan (Grant No. NSTC-114-2410-H-003-027-MY3), and the Institute for Research Excellence in Learning Sciences of National Taiwan Normal University (NTNU) from the Featured Areas Research Center Program within the framework of the Higher Education Sprout Project by the Ministry of Education (MOE) in Taiwan.

## 7. Appendix: 2025 CSAT Earth Science I Problem Set

For the reader's reference, the Korean original text has been translated into English and attached as an Appendix. The quality of the translation is not guaranteed.